\documentclass[
]{ceurart}

\usepackage{float}
\usepackage{subfigure}
\usepackage[english]{babel}
\usepackage[utf8]{inputenc}
\usepackage{algorithm}
\usepackage[noend]{algpseudocode}
\usepackage{graphicx}
\usepackage{natbib}
\usepackage{lipsum} % for context

\begin{document}

\copyrightyear{2021}
\copyrightclause{Copyright for this paper by its authors.\\
  Use permitted under Creative Commons License Attribution 4.0
  International (CC BY 4.0).}

\conference{CERC 2021: Collaborative European Research Conference,
  September 09--10, 2021, Cork, Ireland}

\title{Splitfed learning without client-side synchronization: Analyzing client-side split network portion size to overall performance}
\author[1]{Praveen Joshi}
\ead{praveen.joshi@mycit.ie}
\address[1]{            Munster Technological University, Ireland
            }
\author[2]{Chandra Thapa}
\ead{chandra.thapa@data61.csiro.au}
\address[2]{CSIRO Data61, Australia
            }
\author[2]{Seyit Camtepe}
\ead{seyit.camtepe@data61.csiro.au}

\author[1]{Mohammed Hasanuzzaman}
\ead{Mohammed.Hasanuzzaman@mtu.ie}

\author[1]{Ted Scully}
\ead{Ted.Scully@mtu.ie}

\author[1]{Haithem Afli}
\ead{Haithem.afli@mtu.ie}

%===============================================
%           ABSTRACT
\begin{abstract}
 %Machine learning-based models have seen wide adoption amongst the research communities and the industry due to their significant success on non-trivial tasks such as natural language understanding, speech synthesis, and image recognition tasks. 
 %
%  With the proliferation of Internet-of-Things (IoT), we are now generating enormous data every second, and machine learning techniques provide a means to analyze those data. As the network of IoT is becoming complex and data is exponentially growing, analysis in such a scenario requires the distributed machine learning algorithm.  
%  Federated Learning, Split Learning, and SplitFed Learning are three recent developments in distributed machine learning that are gaining attention due to their ability to preserve raw data. Analysis of big data for anomaly detection, large scale image classification, stock market time-series analysis etc., are some of the key applications of the distributed machine learning algorithm.
 %
Federated Learning (FL), Split Learning (SL), and SplitFed Learning (SFL) are three recent developments in distributed machine learning that are gaining attention due to their ability to preserve the privacy of raw data. Thus, they are widely applicable in various domains where data is sensitive, such as large-scale medical image classification, internet-of-medical-things, and cross-organization phishing email detection. 
SFL is developed on the confluence point of FL and SL. It brings the best of  FL and SL by providing parallel client-side machine learning model updates from the FL paradigm and a higher level of model privacy (while training) by splitting the model between the clients and server coming from SL. 
However, SFL has communication and computation overhead at the client-side due to the requirement of client-side model synchronization. For the resource-constrained client-side, removal of such requirements is required to gain efficiency in the learning. In this regard, this paper studies SFL without client-side model synchronization. The resulting architecture is known as \emph{Multi-head Split Learning}. 
%
%In our experiment, we assessed the performance of the model with and without client-side aggregation. Our observations show that results from SplitFed with and without client-side aggregation are comparable (client-side aggregation gives 1-2\% better accuracy on the test set). 
%
Our empirical studies considering the ResNet18 model on MNIST data under IID data distribution among distributed clients find that Multi-head Split Learning is feasible. Its performance is comparable to the SFL. Moreover, SFL provides only 1\%-2\% better accuracy than Multi-head Split Learning on the MNIST test set. 
%
%We can save our computational and communication requirements at the client-side by removing the fed server while developing the comparable model. 
%
%Furthermore, finding the split point in the DNN model split process of SL is challenging because of the trade-off between the computational power held by clients and the data leakage after the split point from the client end. In such a scenario, we have evaluated whether or not model-split impacts the model's overall performance. Our results on uniformly distributed data amongst the clients showed that model performance on different split does not significantly (on MNIST test dataset without client-side aggregation Resnet-18 model reported accuracy within 96.71\%- 97.36\% ) impacts the performance of the model.
%
To further strengthen our results, we study the Multi-head Split Learning with various client-side model portions and its impact on the overall performance. To this end, our results find a minimal impact on the overall performance of the model.

\vskip-10pt
\end{abstract}
% \begin{keywords}
%   Distributed collaborative machine learning \sep
%   Split learning
% \end{keywords}
\maketitle

%===============================================
%           INTRODUCTION
%==============================================
\section{Introduction}
\label{sec:introuction}

In the world of data, the security and privacy of individuals have now become one of the major concerns. To avoid data misuse, several restrictions such as the General Data Protection Regulation (GDPR)~\cite{chassang2017impact}, Personal Data Protection Act (PDP)~\cite{azzi2018challenges}, and Cybersecurity Law of the People's Republic (CLPR) of China~\cite{qi2018assessing} have been introduced. These regulations are strictly practiced making data aggregation from distributed devices and regions almost impossible~\cite{yang2019federated}. To accommodate such restrictions along with the constraints placed by heterogeneous devices, improvised machine learning (ML) approaches were sought.
Federated Learning~\cite{konevcny2016federated} and Split Learning~\cite{poirot2019split} are two such ML approaches that enable safeguarding the raw data and offload computations at the central server by pushing a part of the computation to the end devices. \par 

Federated learning (FL) leverages the distributed resources to train an ML model collaboratively. More precisely, in FL, multiple devices collaboratively offer resources to train the ML model while keeping the raw data to themselves, as in no raw data leaves the place of its origin~\cite{yang2019federated}. 
%However, FL utilizes the resources of distributed devices to train the ML model and ensures the privacy of the individual's raw data.\par
%
%The significant drawbacks of FL are the challenge to train a larger ML model because of the constraint resources available at heterogeneous devices \cite{li2020federated} and obtaining the model privacy as the whole model is always shared with client\cite{thapa2020splitfed}.  
%
The main drawbacks of FL are two folds. Firstly, training a large ML model in resource-constrained end devices is difficult~\cite{li2020federated}. Secondly, all participating end devices and the server has the full trained model. This does not preserve the model privacy while training like in split learning~\cite{thapa2020splitfed}.   

To overcome these drawbacks, Split Learning (SL) enables model split and training the split model portions collaboratively at the client-side and the server-side separately~\cite{singh2019detailed}. The clients and the server never have access to the model updates (gradients) of each other's model portion once the training starts. 
%
%provides a way in which the initial layers are trained in the client-side, and the rest of the layers reside inside the server \cite{singh2019detailed}. 
%
%The model split mitigates two concerns. Firstly, raw data never leaves its origin; secondly, SL can efficiently train a larger ML model as the computational task can be offloaded to the server without sharing the raw data. 
%
This way, SL enables training large models in an environment with low-end devices such as internet-of-things and preserves the model's privacy while training. Also, it keeps the raw data to its origin (the analyst has no access to the raw data at all times).
However, at a time, SL considers only one client and the server while training. This forces other clients to be idle and wait for their turn to train with the server~\cite{thapa2020splitfed}. \par 
%and there is a need to  at a time in SL the sequential nature of SL trainingmakes it slower as compare to FL in terms of model convergence . \par

To mitigate the drawback of FL having a lower level of model privacy while training and the inability of SL to train the ML model in parallel, specifically among the clients, the SplitFed learning (SFL) is recently proposed~\cite{thapa2020splitfed, splitbook}. SFL combines the best of the FL and SL. In this approach, an ML model is split between the client and the server (like in SL). In contrast to SL, multiple identical split of ML model, i.e., the client-side model portion, is shared across the clients. The server-side model portion is provided to the server. In each forward pass, all clients perform the forward propagation in parallel and independently. Then the activation vectors of the end layer (client-side model portion) are passed to the server. The server then processes the forward and backpropagation for its server-side model on the activation vectors. In backpropagation, the server returns the respective gradients of their activation vectors to the clients. Afterward, each client performs the backpropagation on the gradients they received from the server. After each forward and backward pass, all client-side models and server-side models aggregate their weights and form the one global model, specifically in SplitFedV1. The aggregation is done independently at the client-side (by using fed server) and server-side. 
In another version of the SFL called SplitFedV2, the authors changed the training setting for the server-side model. Instead of aggregating the server-side model at each epoch, the server keeps training one server-side model with the activation vectors from all the clients.  
%
%training 'n' server-side model for the 'n' number of clients, only one common server-side model was trained. For this variant, client-side forward and backward propagation with client-side aggregation remains the same as SplitFedV1   \par

Despite the improvements in SFL, model synchronization is needed at the client-side that is obtained through model aggregation and sharing. This is done to make the global model (joint client-side model and server-side model) consistent at the end of each epoch. However, the model synchronization brings the computation and communication overhead at the client-side. This would be significant if the number of clients grows significantly.
%Client-side model aggregation helps the global model to be consistent with clients after each epoch. Although, a significant computational and communication overhead increases while aggregating the client-side model weights, which raises with the number of clients collaborating in the model training.This research is carried to find the effectiveness of the SFL model without client-side model aggregation.  Based on the motivation, we answer the following two research questions in this paper:
%
In this regard, this paper studies the SFL without client-side model synchronization. The resulting model architecture is called \emph{Multi-head Split Learning} (MHSL). We summarize our contributions under two research questions stated in the following:

\subsection{Our contributions}
\begin{itemize}
    \item [\textbf{RQ1}] Can we allow splitfed learning without client-side model synchronization?\par
    We study the feasibility of MHSL. Our empirical studies on IID distributed MNIST and CIFAR-10 data among five clients find a similar result in MHSL and SFL. Moreover, SFL is slightly (1\%-2\%) better than MHSL on the MNIST. 
    %
    %For CIFAR-10, SFL and MHSL are still converging and requires more epochs before any deduction can be made.
    For CIFAR-10, SFL is better by around 10\% than MHSL at the 20 global epoch. However, both SFL and MHSL performance is below 60\% (low), thus requires further studies to make any conclusion.  
    
    %comparable results to the architecture with the client-side aggregation for the MNIST dataset. For the CIFAR-10 dataset, architecture was converging at a slower speed without client-side aggregation compared to counterpart as observer till 20 epochs.
    
    \item [\textbf{RQ2}] Is there any effect on the overall performance if we change the number of layers at the client-side model portions?\par
    Performance of SFL and MHSL under different combinations of layers dispersed at the client-side, and the server-side behaved identically. No significant deviation in model convergence and their performance are observed for any of the client-side and the server-side model's combinations in our experiments.
\end{itemize}

%===============================================
%           Experiment setup
%==============================================
\section{Experiment setup}
\label{sec:setup}

\begin{figure}[]
\centering 
\setlength
\fboxsep{0pt} 
\setlength\fboxrule{0.25pt} 
\fbox{\includegraphics[width=3.5in]{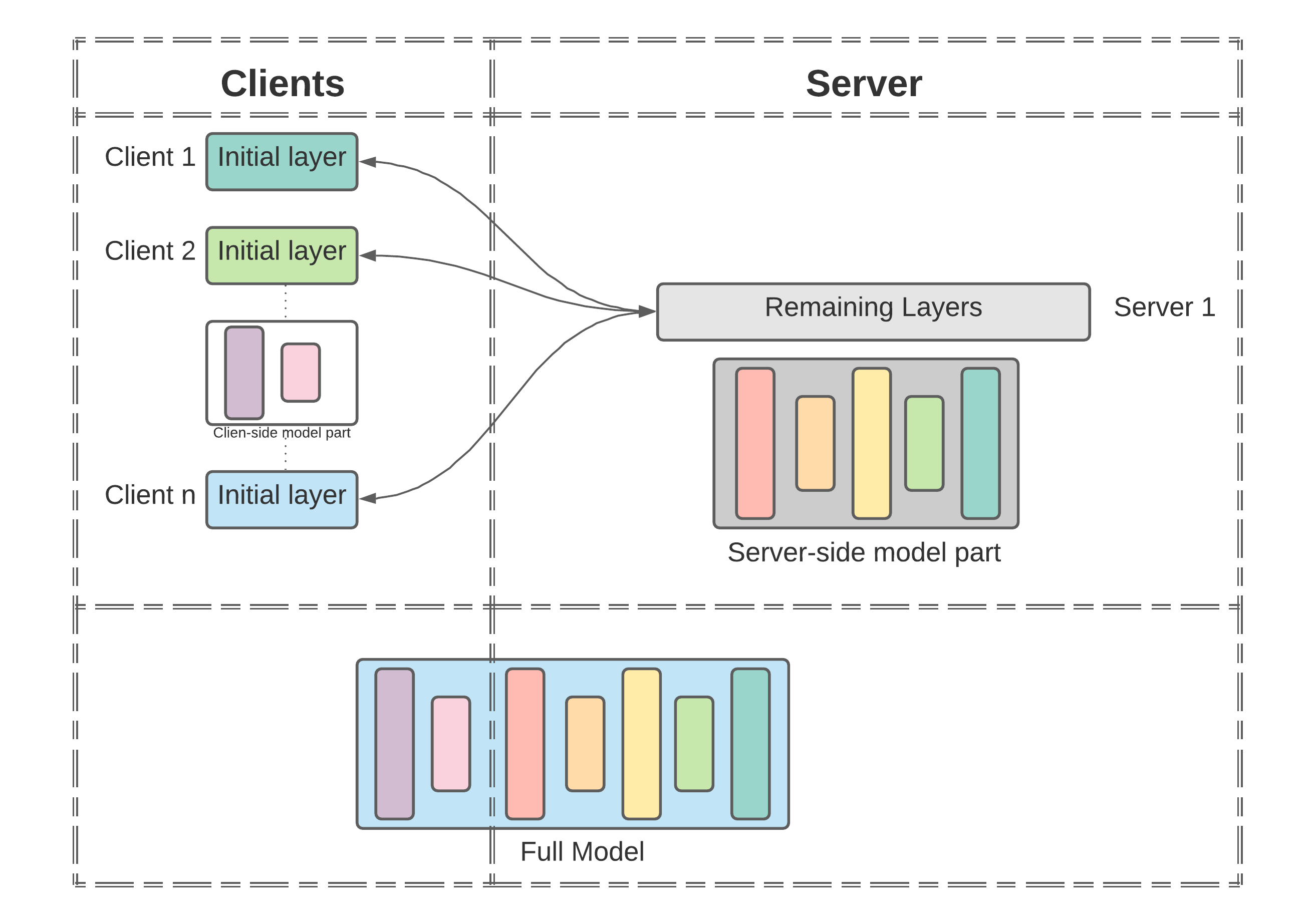}}
\caption{Multi-head split learning architecture.} 
\label{fig:SplitfedV2WCA} 
\vskip-15pt
\end{figure}

%The experiment setup is inspired by the SplitFedV2 architecture because of the less communication and computation requirements as compared to SplitFedV1 \cite{thapa2020splitfed}. 
For the experiment purpose, we choose SplitFedV2 in this paper. This makes our analysis more focused on the split learning side. Moreover, we study if the federated learning part can be removed from the SFL, resulting in Multi-head Split Learning (MHSL).  
%
%However, in experimental setup, client-side aggregation is not considered to remove the additional computation and communication required for the federated aggregation at the client-side (Multi-head split learning). 
%
The overall architecture of MHSL is depicted in Figure~\ref{fig:SplitfedV2WCA}. The model $W$ is split into two portions; client-side model $W_c$ portion and server-side model $W_s$ portion. For the clients, their models are represented by $W_c^i$, where $i\in \{1,2,\dotsc, N\}$ is the client's label. The global model $W$ is formed by concatenating the $W_c$ and $W_s$, i.e., $[W_c W_s]$ once the training completes. \par 

%For experimental setup of the multi-head split learning, refer to figure \ref{fig:SplitfedV2WCA}. It shows the interaction between the client-side model ( \textbf{'Wc\{i\}'} being the client-side model weight of 'i'th client) and the server-side model (\textbf{'Ws'} being the server-side model weight). Therefore global model can be achieved by combining the \textbf{Wc\{i\}} and \textbf{Ws}. 
\paragraph{\textbf{How the final full model is formed in Multi-head Split Learning?}} Unlike SFL, MHSL removes the fed server and the synchronization of $W_c^i$ at the end of each epoch. During the whole training, $W_c^i$ are trained independently by their clients with the server. But, at the end of the whole training, the global full model $W$ is constructed from any one $W_c^i$ and concatenating it with $W_s$. To enable this way of constructing the final trained model, we keep the test data the same over all clients and only keep the training data localized. Thus, if the test results for all clients are similar, then it is reasonable to pick any $W_c^i$ for the final full model.   

%Unlike SFL, FL is not used to aggregate the client-side model. For each epoch, in random without replacement, a client is selected, then performs the forward pass on the local dataset and passes the smashed data to the central server. The central server then carries out forward propagation with the smashed data received from the client. Next, the central server computes the loss and performs the backpropagation on the server-side model to update its weight. Finally, gradients against the smashed data concerning the loss are sent back to the client, which then calculates the respective gradients and updates the client-side model. Thus, for each global epoch, all the clients participate precisely once. Each of the clients evaluates the test accuracy on the global test dataset instead of the localized dataset. The global test dataset for evaluation was taken to check that all the client-side models are approximately the same at any point in time. 

Our program is written using python 3.7.6 and PyTorch 1.2.0 library. The experiments are conducted in a system having a Tesla P100-PCI-E-16GB GPU machine. We observe the training and testing loss and accuracy at each global epoch (once the server trains with all the activation vectors received from all clients). We consider the client-level performance. All the clients were selected to participate at least once at a global epoch without repetition for the current setup.

\subsection{Dataset}
\label{sec:Dataset}
For our experiments, two widely used image datasets, namely, MNIST and CIFAR-10, are selected. Moreover, this dataset maintains the closeness of our results with the reported results in the original paper SplitFedV2. MNIST~\cite{lecun1998mnist} dataset consists of 60,000 images in the training dataset and 10,000 images in the test dataset. The dimension of each of the images in the MNIST dataset is 784 ($28\times28$) in grayscale. Another dataset used for experimentation is CIFAR-10 \cite{krizhevsky2009learning}, consisting of 50,000 images in the training set and 10,000 images in the test dataset. Each image corresponds to the dimension of 3072 ($32\times32$). For the summary, refer to Table~\ref{tab:dataset}. Both of the datasets have ten classes for prediction. For the experimentation, color random horizontal flipping, random rotation, normalization, and cropping on MNIST and CIFAR-10 are conducted to avoid the problem of over-fitting. In addition, for all our experiments, data is assumed to be uniformly and identically distributed amongst five clients.

\begin{table}
\caption{\label{tab:dataset}Datasets used in our experiment setup.}
\footnotesize
\centering
\begin{tabular}{cccc}
\hline Dataset & \multicolumn{2}{c} { Training samples Testing samples } & Image size \\
\hline MNIST & 60,000 & 10,000 & $28 \times 28$ \\
\hline CIFAR-10 & 50,000 & 10,000 & $32 \times 32$ \\
\hline
\label{tab:dataset}
\end{tabular}
\vspace{-0.5cm}
\end{table}

\subsection{Models}
ResNet-18~\cite{he2016deep} network architecture is used for the primary experimentation on the MNIST and CIFAR-10 datasets. The ResNet-18 network was selected because of the discrete ``blocks" structure in every layer of the architecture~\cite{he2016deep}, and it is a standard model for image processing. Resnet-18 blocks were used to split the Resnet-18 between the clients and server to form the client-side and server-side models. Each block performs an operation; an operation in block refers to passing an image through a convolution, batch normalization, and a ReLU activation excluding the last operation in the block. Resnet-18 in the experiment is initialized with a learning rate of 1e-4, and the mini-batch size of BN was set to 64 based on the initial experimentation~\ref{subsec:baseline}. In addition, the first convolutional layer kernel size was set to 7x7, remaining convolutional layers used 3x3 kernels as described in the model architecture Table~\ref{tab:model}.

\begin{table}
\caption{\label{tab:model}Model Architecture used in the experimental setup.}
\footnotesize
\centering
\begin{tabular}{cccc}
\hline Architecture & No. of parameters & Layers & Kernel size \\
\hline ResNet18~\cite{he2016deep} & $11.7$ million & 18 & $(7 \times 7),(3 \times 3)$ \\
\hline
\label{tab:model}
\end{tabular}
\vspace{-0.5cm}
\end{table}

%===============================================
%           RESULTS
%==============================================
\section{Results}
\label{sec:result}
This section presents the empirical results on the MNIST and CIFAR-10 datasets. The results are divided into three parts. First, section~\ref{subsec:baseline} offers results obtained while training the centralized version of the Resnet-18 on the CIFAR-10 and MNIST datasets. In this section~\ref{subsec:baseline}, we compare the results of SplitFedV2 and MHSL on MNIST and CIFAR-10 datasets. For both datasets, we consider five clients to have comparable results, as shown in SplitFedv2 research~\cite{thapa2020splitfed}. In both the architecture, we have kept the initial layer inside the clients (as a client-side model portion), and the rest of the layers reside in the server (as a server-side model portion). Finally, in section~\ref{subsec:rq2}, we have presented our empirical results indicating the impact of the model split on the overall performance of the ResNet-18 model. 

\subsection{Baseline result}
\label{subsec:baseline}
\begin{figure}[]
\centering
\setlength
\fboxsep{0pt} 
\setlength\fboxrule{0.25pt} 
	\subfigure[]{
		\fbox{\includegraphics[width=0.45\columnwidth]{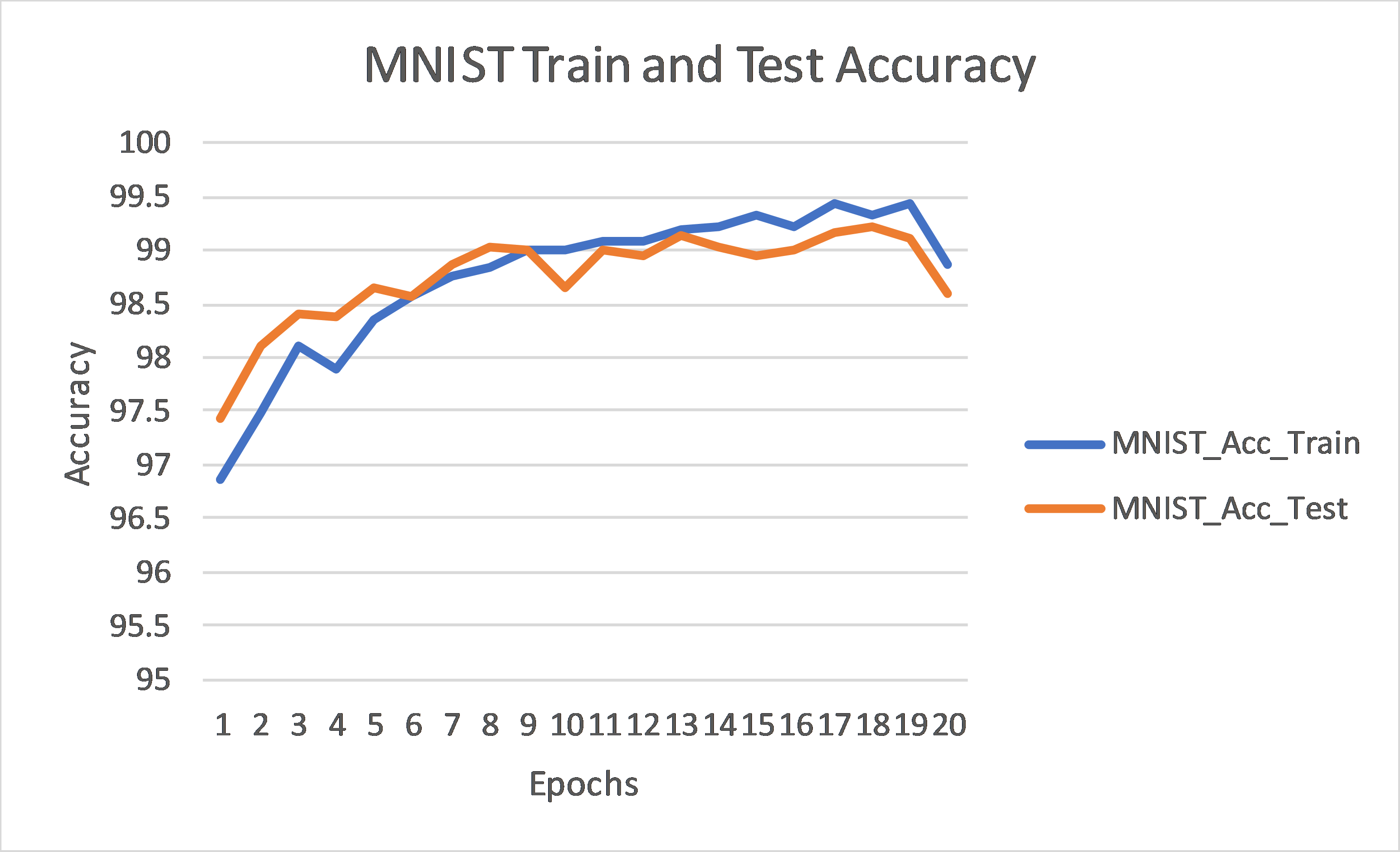}}
	}
	\hskip2pt
	\subfigure[]{
		\fbox{\includegraphics[width=0.45\columnwidth]{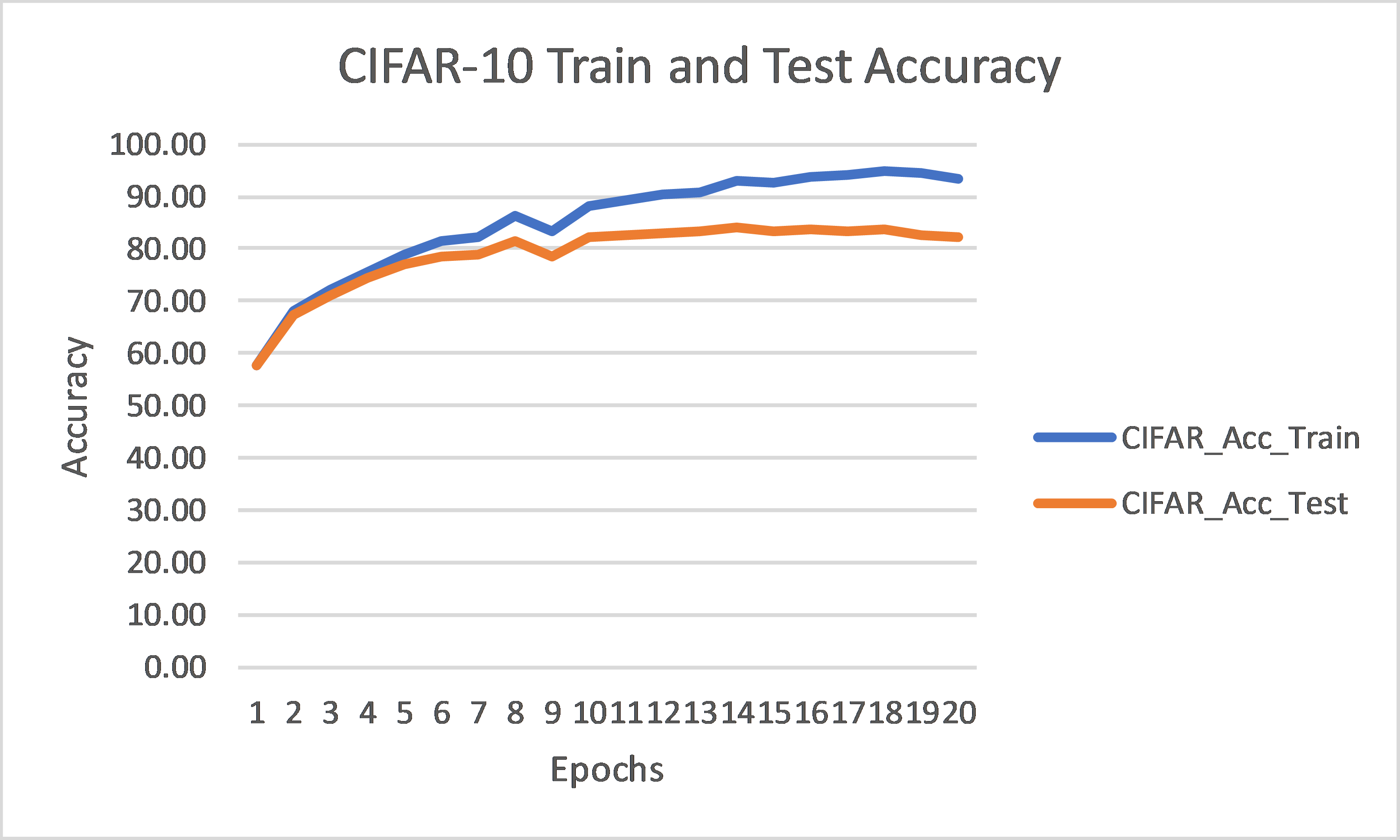}}
	}
	\caption{Train and test Accuracy of ResNet18 model on MNIST and CIFAR-10 in the centralized training.}
	\label{fig:baseline}
\end{figure}

For the baseline, MNIST and CIFAR-10 are subjected to ResNet-18 model architecture. For both the datasets, data-augmentation techniques are the same as discussed in the section~\ref{sec:Dataset}. Training of the ResNet-18 model is done in a centralized manner, i.e., the whole model resided in the server without any split, and all data are available to the server. The convergence curves of both the train and test accuracies for both datasets are shown in Figure~\ref{fig:baseline}.

\subsection{Experiment1: Corresponding to \textbf{RQ1}}
\label{subsec:rq1}
\begin{figure}[]
\centering
\setlength
\fboxsep{0pt} 
\setlength\fboxrule{0.25pt} 
	\subfigure[]{
		\fbox{\includegraphics[width=0.45\columnwidth]{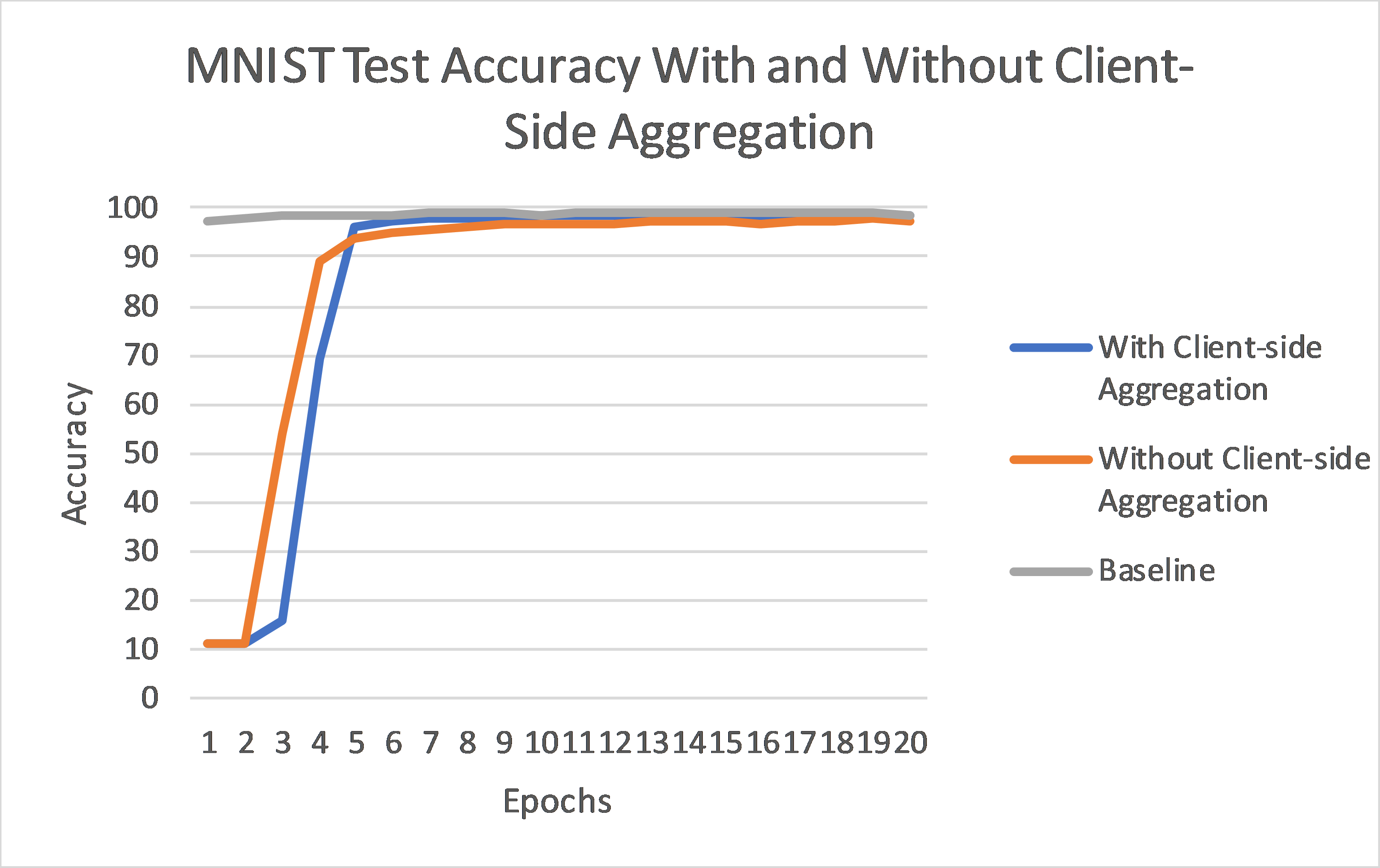}}
	}
	\hskip2pt
	\subfigure[]{
		\fbox{\includegraphics[width=0.45\columnwidth]{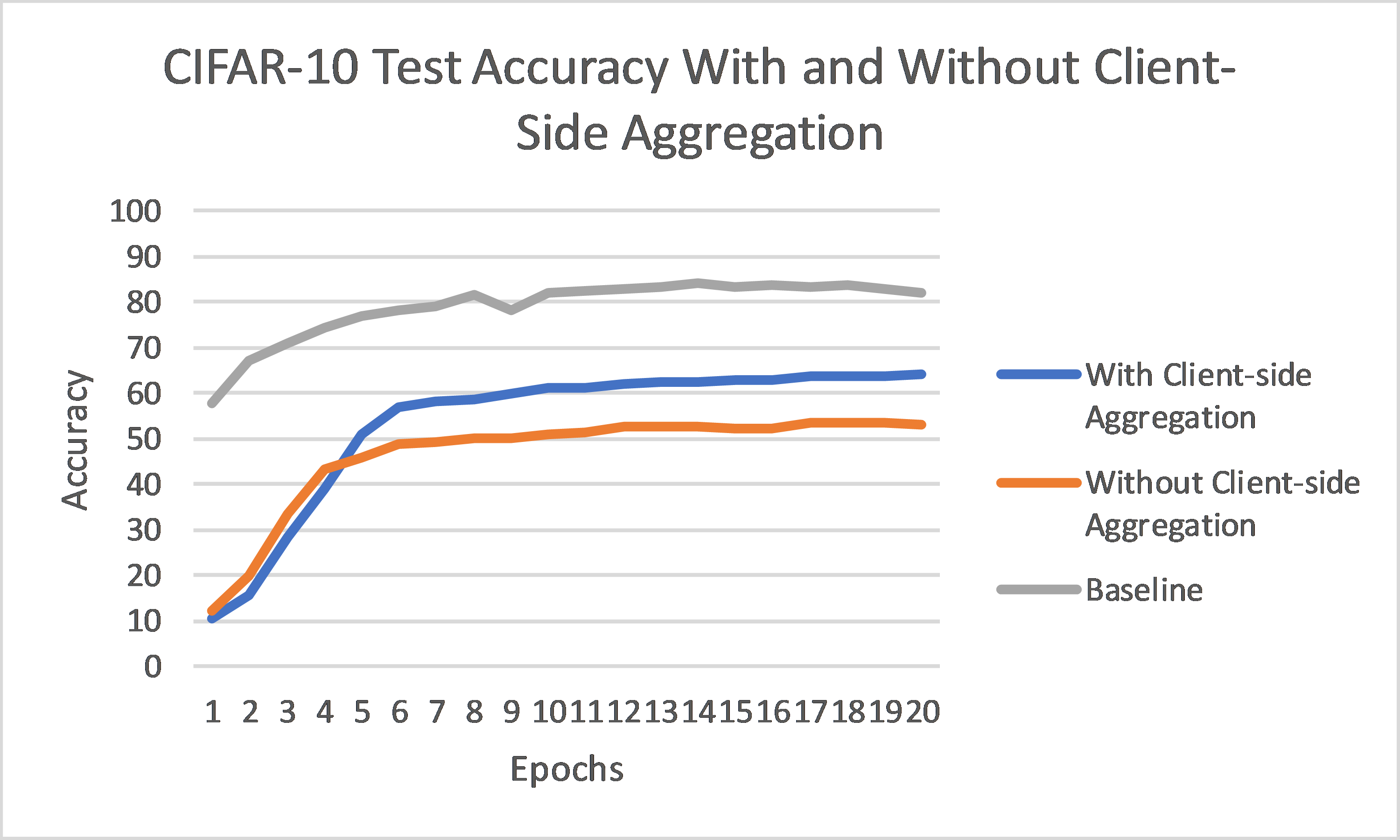}}
	}
	\caption{Test accuracy with client side aggregation (i.e., SFL) and without client-side aggregation (i.e., MHSL) on (a) MNIST and (b) CIFAR-10.}
	\label{fig:rq1}
\end{figure}

This section evaluated the impact of client-side aggregation by splitting the model on the first layer. The very first layer reside at the client-side (client-side model portion) and the remaining on the server-side (server-side model portion). Experimental results in terms of test accuracy on MNIST and CIFAR-10 dataset with and without client-side aggregation are shown in Figure~\ref{fig:rq1}.

From the results in Figure~\ref{fig:rq1}(a), it is evident that results are similar for SFL and MHSL. For CIFAR-10, the performance for both SFL and MHSL are quite lower than the baseline, but the result is better in the case of MNIST.

%from ML model with client-side aggregation in the MNIST dataset is comparable to ML model without client-side aggregation. Although for the CIFAR-10 dataset in both, the ML model is still converging at epoch 20 as shown in figure \ref{fig:rq1} (b).

\subsection{Experiment2: Corresponding to RQ2}
\label{subsec:rq2}
\begin{figure}[]
\centering
\setlength
\fboxsep{0pt} 
\setlength\fboxrule{0.25pt} 
	\subfigure[]{
		\fbox{\includegraphics[width=0.45\columnwidth]{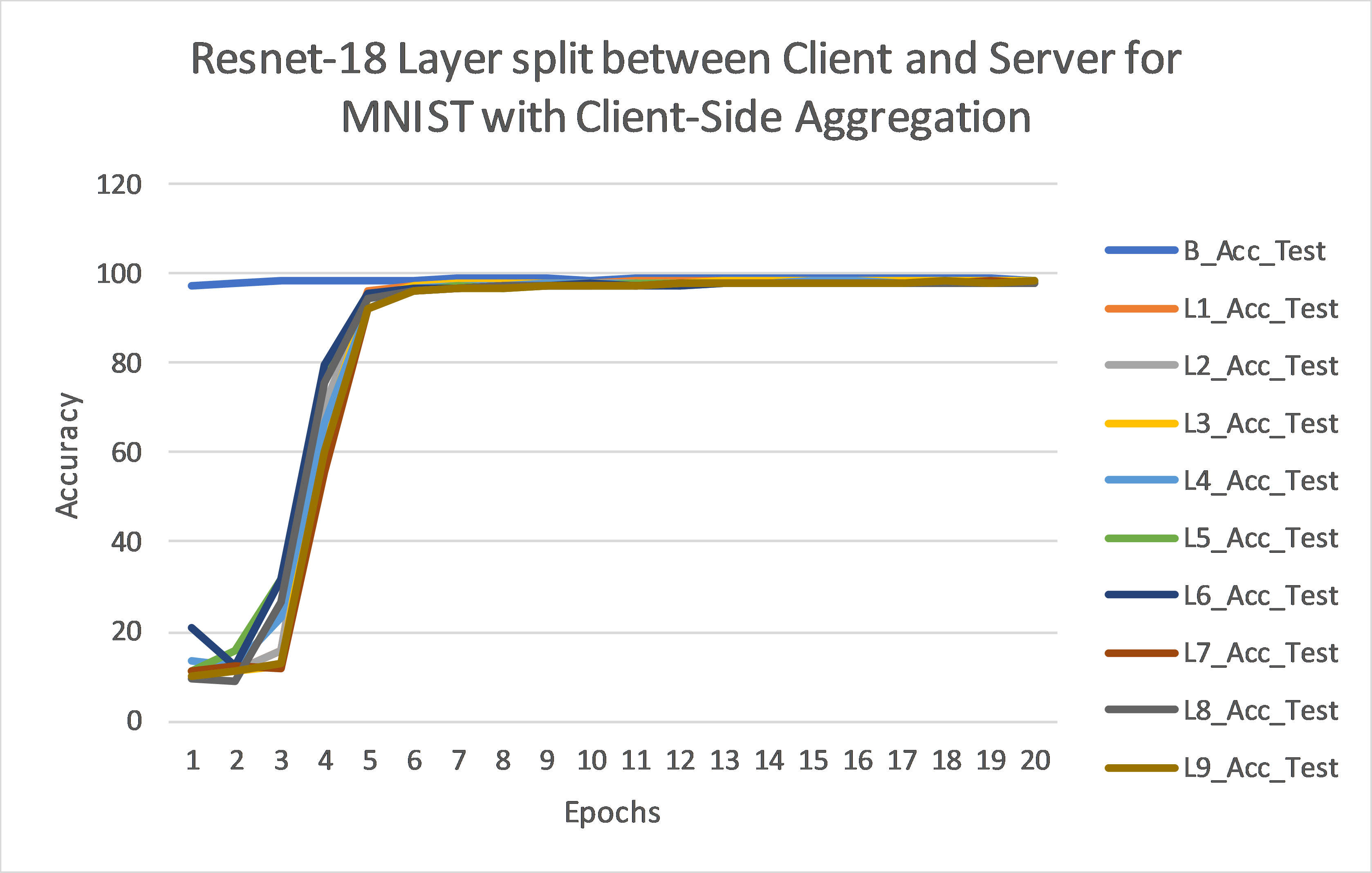}}
	}
	\hskip2pt
	\subfigure[]{
		\fbox{\includegraphics[width=0.45\columnwidth]{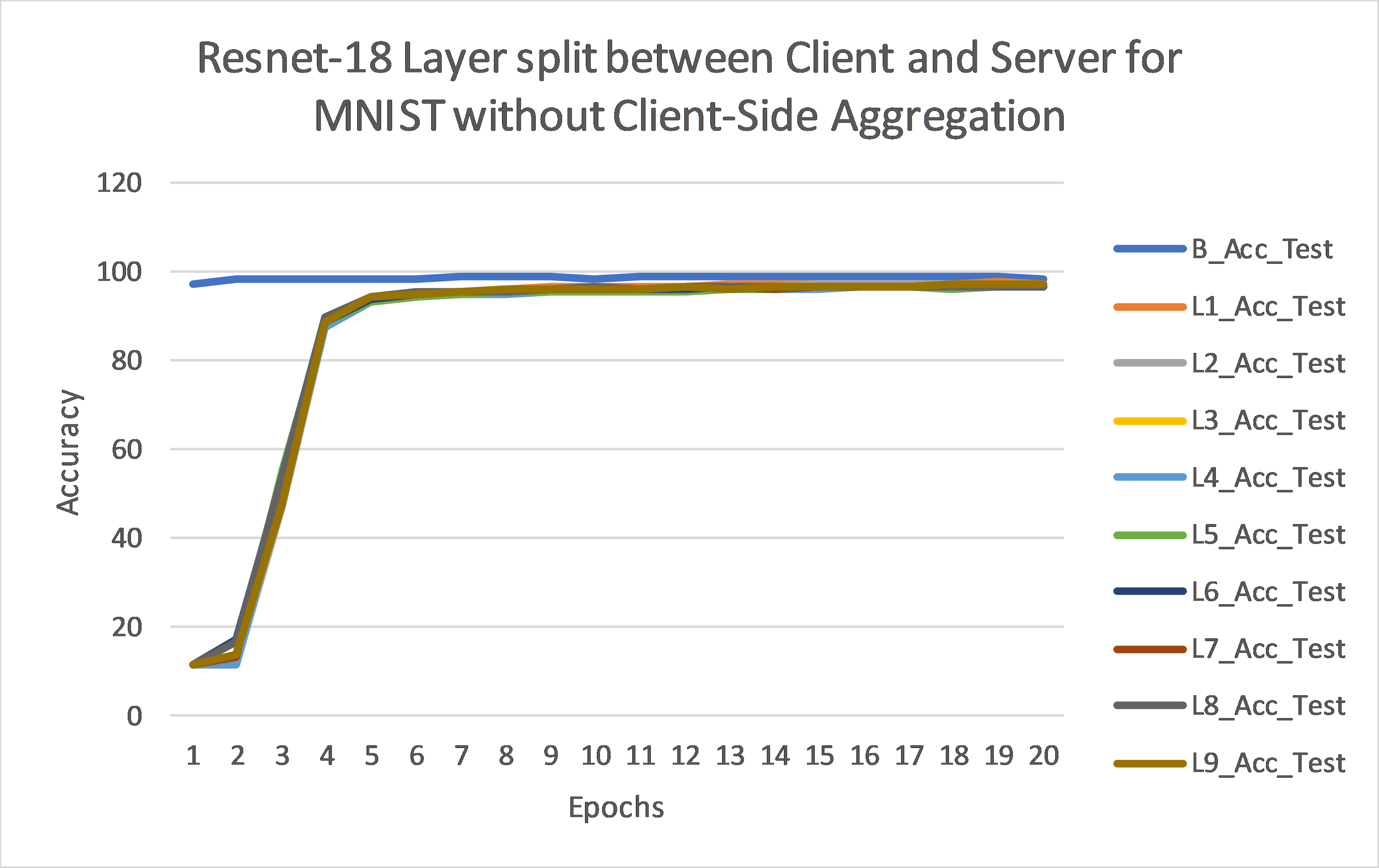}}
	}
	\caption{Test accuracy of ResNet-18 on MNIST (a) with client-side aggregation (i.e., SFL) and (b) without client-side aggregation (i.e., MHSL).}
	\label{fig:rq2}
\end{figure}

This section evaluated the impact of the model split on the overall performance. Test accuracy on MNIST is shown in Figure~\ref{fig:rq2}. 

% \usepackage{graphicx}
\begin{table}[]
\caption{Test Accuracy of ResNet-18 with the model split on different layers.}
\label{tab:rq2_l}
\resizebox{\textwidth}{!}{%
\begin{tabular}{|l|r|r|r|r|r|r|r|r|r|}
\hline
\textbf{Split at layer} &
  \multicolumn{1}{l|}{\textbf{L1}} &
  \multicolumn{1}{l|}{\textbf{L2}} &
  \multicolumn{1}{l|}{\textbf{L3}} &
  \multicolumn{1}{l|}{\textbf{L4}} &
  \multicolumn{1}{l|}{\textbf{L5}} &
  \multicolumn{1}{l|}{\textbf{L6}} &
  \multicolumn{1}{l|}{\textbf{L7}} &
  \multicolumn{1}{l|}{\textbf{L8}} &
  \multicolumn{1}{l|}{\textbf{L9}} \\ \hline
\textbf{Model with Client-Side Aggregation} &
  98.54 &
  98.46 &
  98.56 &
  98.54 &
  98.37 &
  98.21 &
  97.84 &
  98.13 &
  98.25 \\ \hline
\textbf{Model without Client-Side Aggregation} &
  97.23 &
  97.36 &
  96.98 &
  96.71 &
  96.79 &
  96.92 &
  96.93 &
  96.95 &
  97.19 \\ \hline
\end{tabular}%
}
\end{table}

From Table~\ref{tab:rq2_l}, it is evident that SFL and MHSL show a comparable test performance. Overall, our empirical results (both under \textbf{RQ1} and \textbf{RQ2} demonstrate that Multi-head Split Learning (MHSL) is feasible, and there is no significant impact on the performance due to the model split at the various layers of the ResNet-18 model.

%Observed experimental results for \textbf{RQ1} demonstrate the multi-head split learning capability to achieve comparable test accuracy as SplitFedV2 at a reduced cost. Also, \textbf{RQ2} shows that splitting the model at any layer does not impact the test accuracy.

%===============================================
%           CONCLUSION         
%==============================================
\section{Conclusion and future works}
\label{sec:conclusion}
This paper studied SplitFed Learning (SFL) without client-side model synchronization called Multi-head Split Learning (MHSL). 
%
%This paper studied the impact on model performance if client-side aggregation is not performed in a multi-client split learning scenario. 
%
Our experiments with ResNet-18 on the MNIST dataset demonstrated that MHSL is feasible. In other words, our studies suggested that the fed server and the client-side model synchronization can be removed from SFL to reduce the communication and computation overhead at the client side.  
%
%We observed values of the ResNet-18 networks on the MNIST and CIFAR-10 datasets for our experiment. Experimental results demonstrated that removing the client-side aggregation ML model can achieve comparable results to the ML model with client-side aggregation. Furthermore, this setup reduces the requirement of an additional server for performing FL for client-side model aggregation. 
%
In addition, our experiments with different combinations of model portion size at the client-side and the server-side found a negligible effect on the overall performance. This suggests the possibility of dynamic allocation of layers to the clients based on the computation power without any significant loss in the model performance. 

This paper is the first step to find the feasibility of MHSL and the effect of the split network portion sizes to the overall performance. In the future, it will be interesting to see more exhaustive experiments and theoretical analysis on the convergence guarantee with the different models, various datasets, and under a larger number of clients in the experimental setup. Also, experimenting with the setup for non-IID data setup will be another research direction that can be explored.

%===============================================
%           ACKNOWLEDGEMENTS
%==============================================
\begin{acknowledgments}
This work was supported in part by the  financial  support  of ADVANCE CRT PHD Cohort under Grant Agreement No.18/CRT/6222  and  at  the  ADAPT  SFI  Research  Centre  at Munster Technological University. The ADAPT SFI Centre for Digital Media Technology is funded by Science Foundation Ireland through the SFI Research Centres Programme and is co-funded under the European Regional Development Fund (ERDF) through Grant 13/RC/2106.
\end{acknowledgments}

%===============================================
%           APPENDICES
%==============================================
%\section{Appendices}

{\footnotesize
\bibliography{sample-ceur}}

\appendix

%===============================================
%           APPENDIX SECTION
%==============================================
% \section{Appendix}


\end{document}